\documentclass{article}

 \usepackage[preprint]{neurips_2020}

\usepackage[utf8]{inputenc} 
\usepackage[T1]{fontenc}    
\usepackage{url}            
\usepackage{booktabs}       
\usepackage{amsfonts}       
\usepackage{nicefrac}       
\usepackage{microtype}      
\usepackage{amsmath, amssymb, amsthm, graphicx,color}
\usepackage[colorlinks,
	linkcolor=blue,
	citecolor=blue,
	urlcolor=magenta,
	linktocpage,
	plainpages=false]{hyperref}

\title{Averaging Atmospheric Gas Concentration Data using Wasserstein Barycenters}

\author{Mathieu Barr\'e\\
INRIA \& CS Dept.\\
Ecole Normale Sup\'erieure,\\
PSL Research University\\
Paris, France.
\And
Cl\'ement Giron, Matthieu Mazzolini\\
Kayrros SAS\\
Paris, France.
\And
Alexandre d'Aspremont\\
CNRS \& CS Dept.\\
Ecole Normale Sup\'erieure,\\
PSL Research University\\
Paris, France.
}

\begin{document}

\newcommand{\MB}[1]{{{\color{red}({\bf{Comment MB:}} #1)}}}
\newcommand{\N}{\mathbb{N}}
\newcommand{\meth}{$\mathrm{CH}_4$}
\newcommand{\mty}{Tg \meth/y }
\definecolor{ddarkbrown}{rgb}{0.5,0.2,0.05} \definecolor{bbluegray}{rgb}{0.05,0,0.5}

\renewenvironment{proof}{\textbf{Proof.}}{\QED\bigskip}

\newcommand{\BEAS}{\begin{eqnarray*}}
\newcommand{\EEAS}{\end{eqnarray*}}
\newcommand{\BEA}{\begin{eqnarray}}
\newcommand{\EEA}{\end{eqnarray}}
\newcommand{\BEQ}{\begin{equation}}
\newcommand{\EEQ}{\end{equation}}
\newcommand{\BIT}{\begin{itemize}}
\newcommand{\EIT}{\end{itemize}}
\newcommand{\BNUM}{\begin{enumerate}}
\newcommand{\ENUM}{\end{enumerate}}

\newcommand{\BA}{\begin{array}}
\newcommand{\EA}{\end{array}}

\newcommand{\refp}[1]{(\ref{#1})}

\newcommand{\cf}{{\it cf.}}
\newcommand{\eg}{{\it e.g.}}
\newcommand{\ie}{{\it i.e.}}
\newcommand{\etc}{{\it etc.}}
\newcommand{\ones}{\mathbf 1}

\newcommand{\reals}{{\mathbb R}}
\newcommand{\sreals}{\scriptsize{\mbox{\bf R}}}
\newcommand{\integers}{{\mbox{\bf Z}}}
\newcommand{\eqbydef}{\mathrel{\stackrel{\Delta}{=}}}
\newcommand{\complex}{{\mbox{\bf C}}}
\newcommand{\symm}{{\mbox{\bf S}}}  

\newcommand{\Span}{\mbox{\textrm{span}}}
\newcommand{\Range}{\mbox{\textrm{range}}}
\newcommand{\nullspace}{{\mathcal N}}
\newcommand{\range}{{\mathcal R}}
\newcommand{\diam}{\mathop{\bf radius}}
\newcommand{\sphere}{{\mathbb S}}
\newcommand{\Nullspace}{\mbox{\textrm{nullspace}}}
\newcommand{\Rank}{\mathop{\bf Rank}}
\newcommand{\NumRank}{\mathop{\bf NumRank}}
\newcommand{\NumCard}{\mathop{\bf NumCard}}
\newcommand{\Card}{\mathop{\bf Card}}
\newcommand{\Tr}{\mathop{\bf Tr}}
\newcommand{\diag}{\mathop{\bf diag}}
\newcommand{\lambdamax}{{\lambda_{\rm max}}}
\newcommand{\lambdamin}{\lambda_{\rm min}}
\newcommand{\idm}{\mathbf{I}}

\newcommand{\Expect}{\textstyle{\bf E}}
\newcommand{\Median}{\textstyle\mathop{\bf M}}
\newcommand{\Prob}{\mathop{\bf Prob}}
\newcommand{\erf}{\mathop{\bf erf}}

\newcommand{\Co}{{\mathop {\bf Co}}}
\newcommand{\co}{{\mathop {\bf Co}}}
\newcommand{\Var}{\mathop{\bf var{}}}
\newcommand{\dist}{\mathop{\bf dist{}}}
\newcommand{\Ltwo}{{\bf L}_2}
\newcommand{\QED}{~~\rule[-1pt]{6pt}{6pt}}\def\qed{\QED}
\newcommand{\approxleq}{\mathrel{\smash{\makebox[0pt][l]{\raisebox{-3.4pt}{\small$\sim$}}}{\raisebox{1.1pt}{$<$}}}}
\newcommand{\argmin}{\mathop{\rm argmin}}
\newcommand{\epi}{\mathop{\bf epi}}
\newcommand{\var}{\mathop{\bf var}}

\newcommand{\vol}{\mathop{\bf vol}}
\newcommand{\Vol}{\mathop{\bf vol}}

\newcommand{\dom}{\mathop{\bf dom}}
\newcommand{\aff}{\mathop{\bf aff}}
\newcommand{\cl}{\mathop{\bf cl}}
\newcommand{\Angle}{\mathop{\bf angle}}
\newcommand{\intr}{\mathop{\bf int}}
\newcommand{\relint}{\mathop{\bf rel int}}
\newcommand{\bd}{\mathop{\bf bd}}
\newcommand{\vect}{\mathop{\bf vec}}
\newcommand{\dsp}{\displaystyle}
\newcommand{\foequal}{\simeq}
\newcommand{\VOL}{{\mbox{\bf vol}}}
\newcommand{\argmax}{\mathop{\rm argmax}}
\newcommand{\xopt}{x^{\rm opt}}

\newcommand{\Xb}{{\mbox{\bf X}}}
\newcommand{\xst}{x^\star}
\newcommand{\varphist}{\varphi^\star}
\newcommand{\lambdast}{\lambda^\star}
\newcommand{\Zst}{Z^\star}
\newcommand{\fstar}{f^\star}
\newcommand{\xstar}{x^\star}
\newcommand{\xc}{x^\star}
\newcommand{\lambdac}{\lambda^\star}
\newcommand{\lambdaopt}{\lambda^{\rm opt}}

\newcommand{\geqK}{\mathrel{\succeq_K}}
\newcommand{\gK}{\mathrel{\succ_K}}
\newcommand{\leqK}{\mathrel{\preceq_K}}
\newcommand{\lK}{\mathrel{\prec_K}}
\newcommand{\geqKst}{\mathrel{\succeq_{K^*}}}
\newcommand{\gKst}{\mathrel{\succ_{K^*}}}
\newcommand{\leqKst}{\mathrel{\preceq_{K^*}}}
\newcommand{\lKst}{\mathrel{\prec_{K^*}}}
\newcommand{\geqL}{\mathrel{\succeq_L}}
\newcommand{\gL}{\mathrel{\succ_L}}
\newcommand{\leqL}{\mathrel{\preceq_L}}
\newcommand{\lL}{\mathrel{\prec_L}}
\newcommand{\geqLst}{\mathrel{\succeq_{L^*}}}
\newcommand{\gLst}{\mathrel{\succ_{L^*}}}
\newcommand{\leqLst}{\mathrel{\preceq_{L^*}}}
\newcommand{\lLst}{\mathrel{\prec_{L^*}}}

\newcommand{\realsp}{\mathbf{R}_+^n}
\newcommand{\intrealsp}{\int_{\mathbf{R}_+^n}}

\maketitle

\begin{abstract}
Hyperspectral satellite images report greenhouse gas concentrations worldwide on a daily basis. While taking simple averages of these images over time produces a rough estimate of relative emission rates, atmospheric transport means that simple averages fail to pinpoint the source of these emissions. We propose using Wasserstein barycenters coupled with weather data to average gas concentration data sets and better concentrate the mass around significant sources.
\end{abstract}

\section{Introduction}
Thanks to lower launch costs and a renewed focus on earth observation, there are now several constellations of satellites monitoring greenhouse gas emissions from sun-synchronous orbits. These satellites, notably Sentinel-5P from the European Union's Copernicus program, provide daily hyperspectral images of the entire globe at a resolution of $5.5 \times 7$ km. While anthropogenic emissions of carbon dioxide are drowned by natural sources in short time windows, this is not the case for at least two other important gases, methane (\meth) a very potent greenhouse gas, and nitrogen dioxide ($\mathrm{NO}_2$), a pollutant. We focus on methane here, as it has a longer lifespan than $\mathrm{NO}_2$ (12 years versus a few hours) hence simple averages of concentration images seriously lack contrast.

Anthropogenic methane emissions come from mostly two sources: the oil and gas industry on one hand, agriculture and waste management on the other. Methane has a lifespan of 12 years, much shorter than $\mathrm{CO}_2$, but a global warming potential (GWP) that is 85 five times higher than that of $\mathrm{CO}_2$, over a 20 years window. This means that mitigating methane emissions can have a very significant short term impact on warming. The latest methane budget \citep{Saun20} estimates anthropogenic methane emissions at about 380 \mty (bottom-up), of which 108 \mty is coming from the oil and gas sector and 227 \mty are attributed to agriculture and waste management. Oil and gas emissions are concentrated in a few dense clusters around shale basins such as the Permian in the US, pipelines and major fields in e.g. Turkmenistan or Algeria. Countries and companies with high operational standards tend to have lower emission rates, and the fact that the list of key emitters is relatively short means that the oil and gas sector is both a low cost and high short term impact greenhouse gas emissions mitigation target.

Here, we solve the Wasserstein barycenter problem defined in \citep{Rabi11,Ague11} on emission data sets. Recent advances in computational optimal transport using simple iterative methods \citep{Sink64,Wils69,Cutu13a,Chiz18a} mean that this task is now feasible at scale on satellite image data sets and we refer the reader to \citep*{Peyr19} for a complete treatment. Using proper averages of concentration images has the power to remove part of the noise and better highlight key emitters while lowering the detection threshold.

Atmospheric transport is modeled by two main components, advection and diffusion. While most optimal transport problems usually come with little structural information on the transport model \citep{Peyr19}, in the case of atmospheric transport we have historical weather records describing wind speed and direction, and some information on turbulence. We can use this information to remove biases in the optimal transportation problems underlying the barycenter computation. We thus adapt the local metric in transportation problems to account for known biases introduced by wind, and handle missing pixels using unbalanced transportation problems.

While getting accurate averaged concentrations is important, our main focus is in fact on attribution. We thus seek to solve an {\em inverse transportation problem}: given partial information on the transportation plan (from weather), we seek to identify emission source locations and flow rates consistent with observations, without solving a much heavier (and data intensive) full inversion problem. Early numerical experiments on Sentinel-5P images, reported below, show that Wasserstein averages are much better spatially correlated with oil \& gas activity than simple averages.

\paragraph{Notations} In what follows, we write $h(\cdot)$ the discrete entropy. For $x,y\in \reals_+^n$, $h(x) =\sum_{i=1}^n x_i\log(x_i)$ (with $0\log(0) = 0$) and we write $\text{KL}(\cdot|\cdot)$ the KL divergence, with $KL(x|y) = \sum_{i\in\text{Supp}(y)}x_i\log\left({x_i}/{y_i} \right)$.

\section{Wasserstein Barycenter of Atmospheric Gas Concentrations}
Consider we are given $N\in \N$ squared images $(g^{(k)})_{k\in[1,N]}$ with each $g^{(k)} \in \reals_+^{n\times n}$ representing gas concentration over time on the same region of earth. Pixels represent the mean methane concentration on small squared regions with same side length. In practice images do not need to be square, but we use this convention here for simplicity. 

We use flattened version of the 2D gas images obtained by running through $g^{(k)}$ row by row. Depending on the context we will refer to $g^{(k)}$ as a 2D image in $\reals^{n\times n}$ or as a 1D vector in $\reals^{n^2}$.

\subsection{Problem formulation}
We focus on the 1D discrete transport problem with entropic regularization \citep{Cutu13a}. Consider the transport problem 
\BEQ \label{eq:OT-discret} 
W^C(\mu,\nu) = \underset{P 1 = \mu , P^T1=\nu}{\min}\; Tr(C^TP) + \lambda h(P) 
\EEQ 
in the variable $P \in \reals_+^{n^2\times n^2}$, given non negative distribution variables $\mu$, $\nu \in \reals_+^{n^2}$, a cost matrix $C\in\reals_+^{n^2 \times n^2}$ and a regularization parameter $\lambda$. The optimal value of this problem measures the effort to move the gas from a state $\mu$ to a state $\nu$. With this formulation the total masses of $\mu$ and $\nu$ have to match. This is not compatible with our problem since methane particles are emitted over time (e.g by leaks), and some are disappearing from our measurements (e.g diffusion below the detection threshold, exiting the studied zone, etc.) and some pixels are often missing (because of water, clouds, etc.).

To compare distribution with different masses, one simple idea used in \citep{Gram15} is to add a new dimension (i.e dummy pixels) to the vector $\mu$ and $\nu$ that will contain the excess of mass of $\mu$ (assuming $1^T\mu \geq 1^T\nu$), then apply the classical transport problem \eqref{eq:OT-discret} to these augmented distributions. Optimal transport between distributions with different masses is often called  unbalanced transport. Unbalanced transport has a nice fluid dynamics interpretation with the introduction of a source term that deals with the creation and deletion of matter during the transport \citep{Chiz18u,Lier18}. In practice here, we use a formulation of unbalanced transportation that relaxes marginal constraints in \eqref{eq:OT-discret}. Given another regularization parameter $\lambda_u$, the problem thus becomes
\BEQ \label{eq:OT-unb-discret} W^C_u(\mu,\nu) = \underset{P \in \reals_+^{n^2\times n^2}}{\min}\; Tr(C^TP) + \lambda h(P) + \lambda_u KL(P 1|\mu) + \lambda_u KL(P^T 1|\nu) \EEQ

We are interested in obtaining an aggregation of a sequence $(g^{(k)})$ of methane distributions recorded daily over some period of time. The transportation cost $W^C_u(\cdot,\cdot)$ can be used as a metric between the gas distributions $(g^{(k)})_{i\in[1,N]}$ and, given a sequence of cost matrix $(C^{(k)})_{k\in[1,N]}$, the Wasserstein barycenter (WB) of the $(g^{(k)})$ is computed as
\BEQ\label{eq:OT-unb-barycenter}\tag{WB}
\bar{g} = {\argmin}\;\sum_{i=k}^N W^{C^{(k)}}_u(g^{(k)},g)
\EEQ
in the variable $g\in \reals_+^{n^2}$. Algorithms to solve this convex optimization problem are described in e.g. \citep{Bena15,Chiz18a}.

\subsection{Choice of local cost function}

The cost matrix $C$ plays an important role in the transport problem. In our setting $C_{ij}$ represents the effort necessary to move a particle from position $X_j = (\lceil \tfrac{j}{n} \rceil,[j\; \text{mod}\;n] +1 )$ to position $X_i = (\lceil \tfrac{i}{n} \rceil,[i\; \text{mod}\;n] +1 )$.

In our experiments we use several types of cost matrix, with $C^{(k)}_{ij} = c(X_i,X_j)$ and $c(\cdot,\cdot)$ defined as
\begin{align}
\text{Euclidean metric :}&&& c(x,y) = \|x-y\|^2 \label{eq:l2}\tag{$L_2$}\\
\text{Wasserstein Fisher Rao metric :} &&&\delta >0,\;c(x,y) = -\log\left(\cos^2\left(\tfrac{\|x-y\|}{2\delta}\wedge\tfrac{\pi}{2}\right) \right) \label{eq:WFR}\tag{WFR}\\
\text{Euclidean + Wind :}&&&t>0,\; c(x,y) = \left(\|x-y\|^2 - t\langle w_k,x-y \rangle\right)_+ \label{eq:l2w}\tag{$L_2+W$}
\end{align}

The choice of \eqref{eq:WFR} is motivated by \citep[Corollary 5.9]{Chiz18u} which links the formulation of \eqref{eq:OT-unb-discret} with a fluid mechanics interpretation of the transport, with creation and deletion of mass. When this cost is used, $\lambda_u$ is set to $1$ and the parameter $\delta$ is used to control mass creation.

In the case of the \eqref{eq:l2w} metric, $w_k \in \reals^2$ is a mean wind vector associated with image $g^{(k)}$. The simple idea behind this cost matrix is that, in the presence of wind, it is easier for a particle to move with the wind than against it. Indeed if the shifting vector $x-y$ from $y$ to $x$ follows the direction of the wind $w_k$ the cost is smaller, and $t$ is a scaling parameter to balance the two terms. 


\section{Experiments}
We used the optimal transport package POT \citep{Flam17pot} to compute the Wasserstein barycenters in our experiments, slightly modified to support the use of multiple cost matrices. 

\subsection{Synthetic Experiments}
In this section we present synthetic experiments to illustrate what can be obtained using Wasserstein barycenters and to emphasis the importance of cost matrices choices. Further intuitive examples of Wasserstein barycenters can be found in \cite{Cutu14}.

\begin{figure}[!ht]
    \centering
    \includegraphics[height=0.22\linewidth]{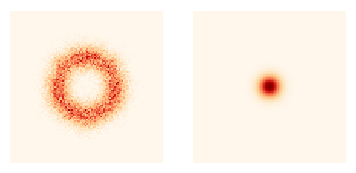}
    \caption{Left: arithmetic mean of $g^{(k)}$. Right: Wasserstein barycenter of  $g^{(k)}$. }
    \label{fig:bar-gaussian}
\end{figure}

First, we look at a simple setting where arithmetic mean and Wasserstein barycenter give very different results presented in Figure~\ref{fig:bar-gaussian}. The $g^{(k)}$ are taken to be Gaussian clouds with unit variance and a mean $\mu_k\in\reals^2$ that rotates around the center of the image with $k$ (to mimic wind changing). We observe that the arithmetic mean described a circle around the center, whereas the Wasserstein barycenter is concentrated on the center. See Appendix~\ref{app:toy-wind} for a synthetic example with \eqref{eq:l2w} cost.

\subsection{Experiments on Real Data}
The data represents $\mathrm{CH}_4$ concentration over a period of time from January 1 2019 until June 1 2020 (see \S~\ref{sec:data} for details). Experiments on the Permian basin are also presented in Appendix~\ref{app:permian}.

\subsubsection{Ohio - West Virginia - Pennsylvania Mines}
The first zone that we look at is a mining area located at the border between Ohio, West Virginia and Pennsylvania. In this region, wind introduces an important bias in the transport of the gas (see Appendix~\ref{app:wind-pen} for more details).


Figure~\ref{fig:pen-bar} shows the result of barycenter computations with different cost matrices. Due to the bias West-East in the wind distribution, we see that the barycenter with costs \eqref{eq:l2w} is shifted to the right compared to barycenter with costs \eqref{eq:l2}.The blue dots correspond to every coal seams that have been exploited at some point in time in the region (see Appendix~\ref{app:source-prod} for sources). While Figure~\ref{fig:pen-bar} only displays coal seams, unconventional gas wells in Southern Pennsylvania are located in the same area \citep{Bark19}. Our Wasserstein barycenter with cost \eqref{eq:l2w} highlights potential methane emitters in Southwestern Pennsylvania. This particular region is well-known for being a significant source of anthropogenic methane \citep{Bark19}, due to underground coal and unconventional gas extraction activities. 
It appears that Wasserstein barycenters are much more correlated with fossil fuels production than classical means.

\begin{figure}[!ht]
    \centering
    \begin{tabular}{ccc}
         \includegraphics[height=0.22\linewidth,width=0.302\linewidth]{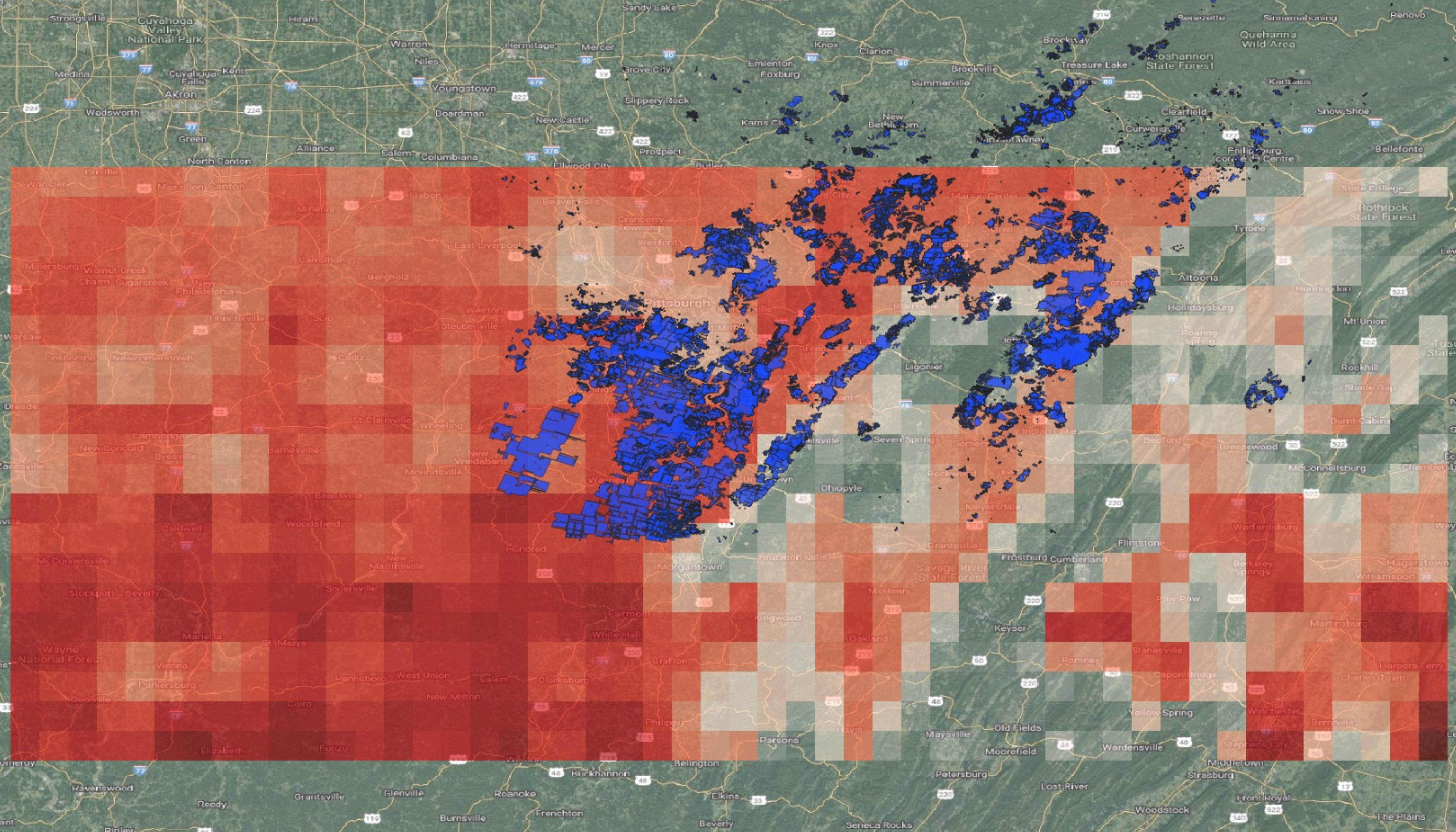}&\includegraphics[height=0.22\linewidth,width=0.302\linewidth]{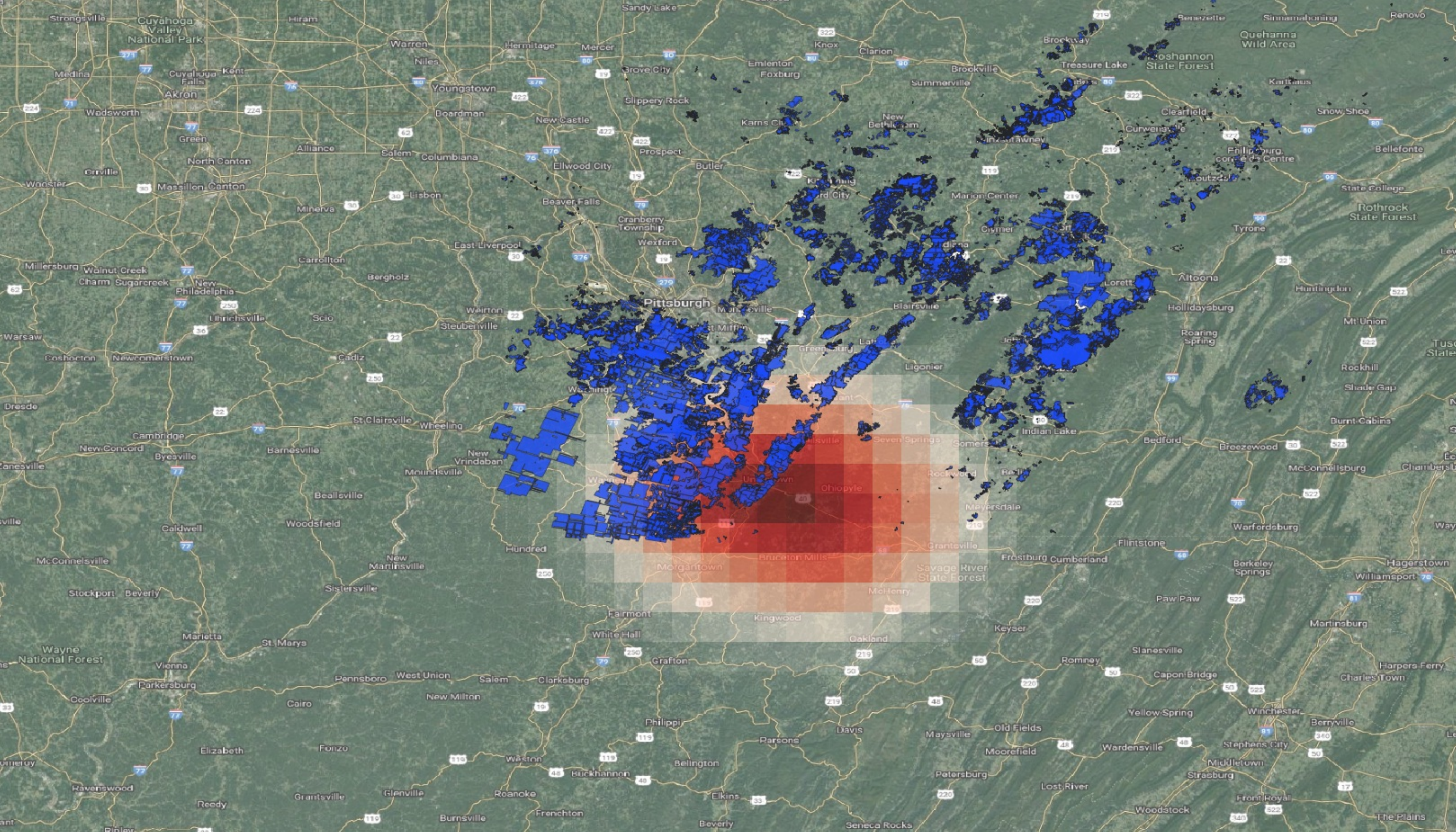}&\includegraphics[height=0.22\linewidth,width=0.302\linewidth]{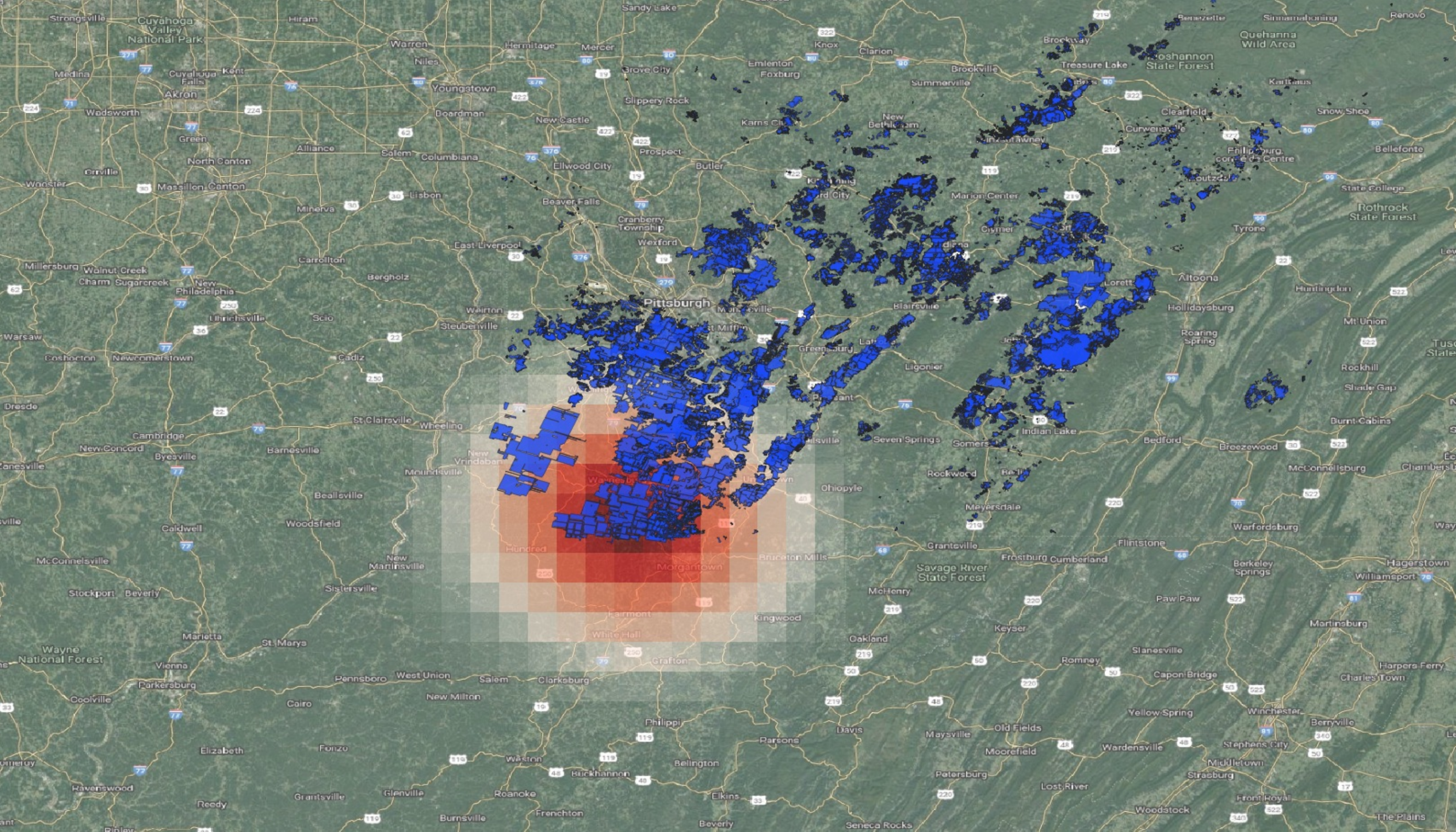}
    \end{tabular}
    \caption{Left: Arithmetic mean. Middle: \eqref{eq:OT-unb-barycenter} + \eqref{eq:l2}. Right: 
    \eqref{eq:OT-unb-barycenter} + \eqref{eq:l2w}}
    \label{fig:pen-bar}
\end{figure}

\subsubsection{Irak-Koweit region}

\begin{figure}[!ht]
    \centering
    \begin{tabular}{ccc}
         \includegraphics[height=0.23\linewidth,width=0.302\linewidth]{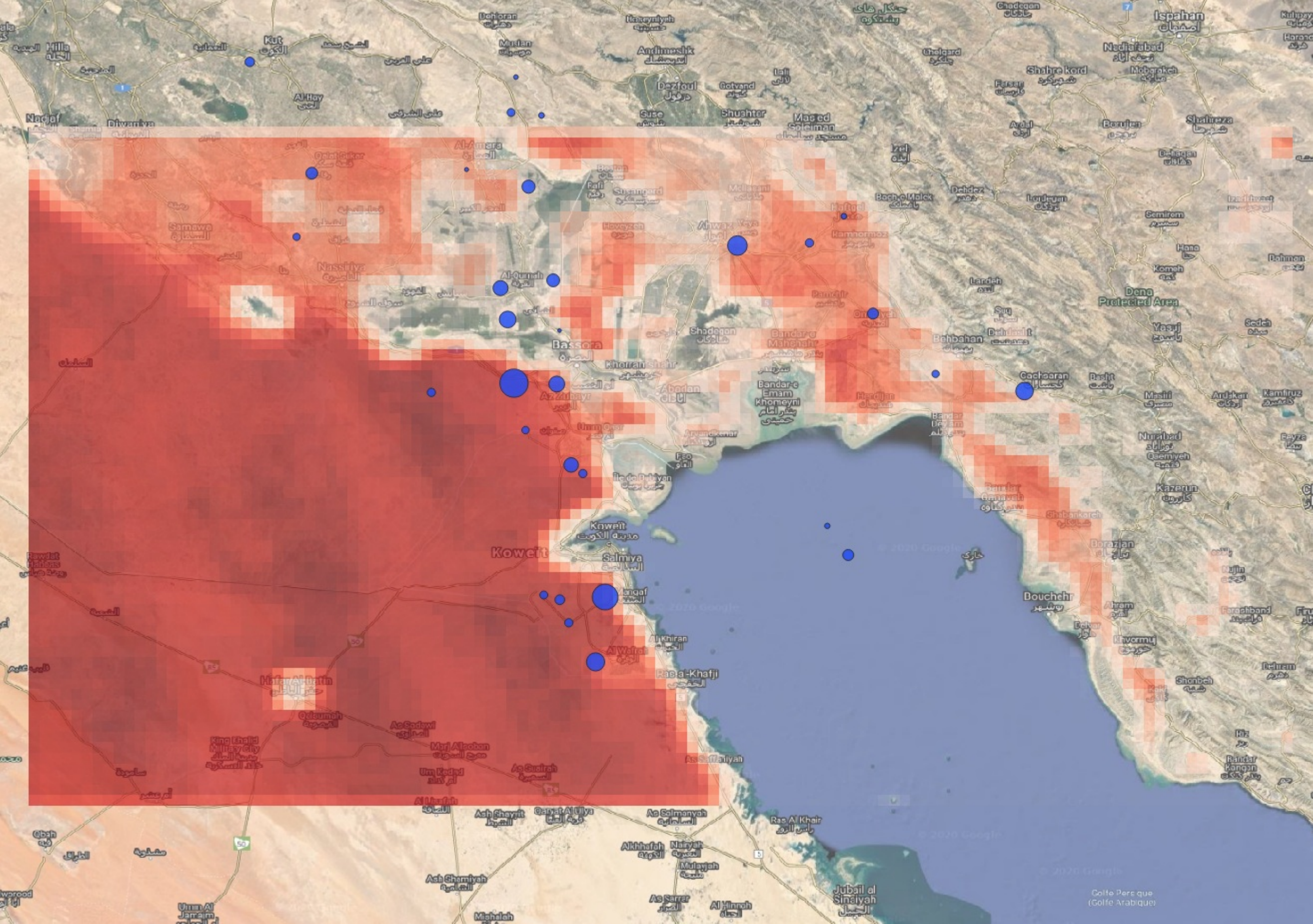}&\includegraphics[height=0.23\linewidth,width=0.302\linewidth]{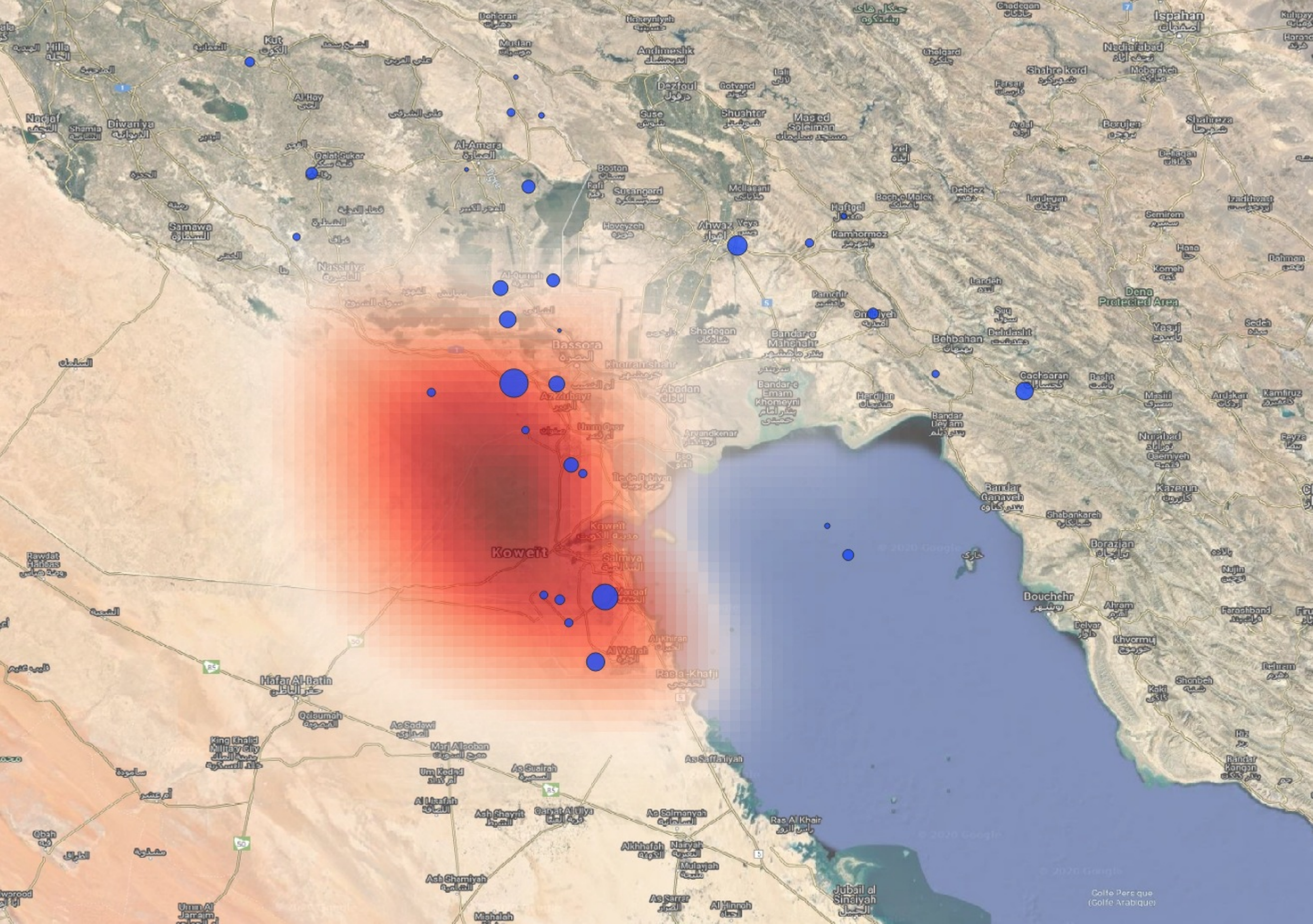}&\includegraphics[height=0.23\linewidth,width=0.302\linewidth]{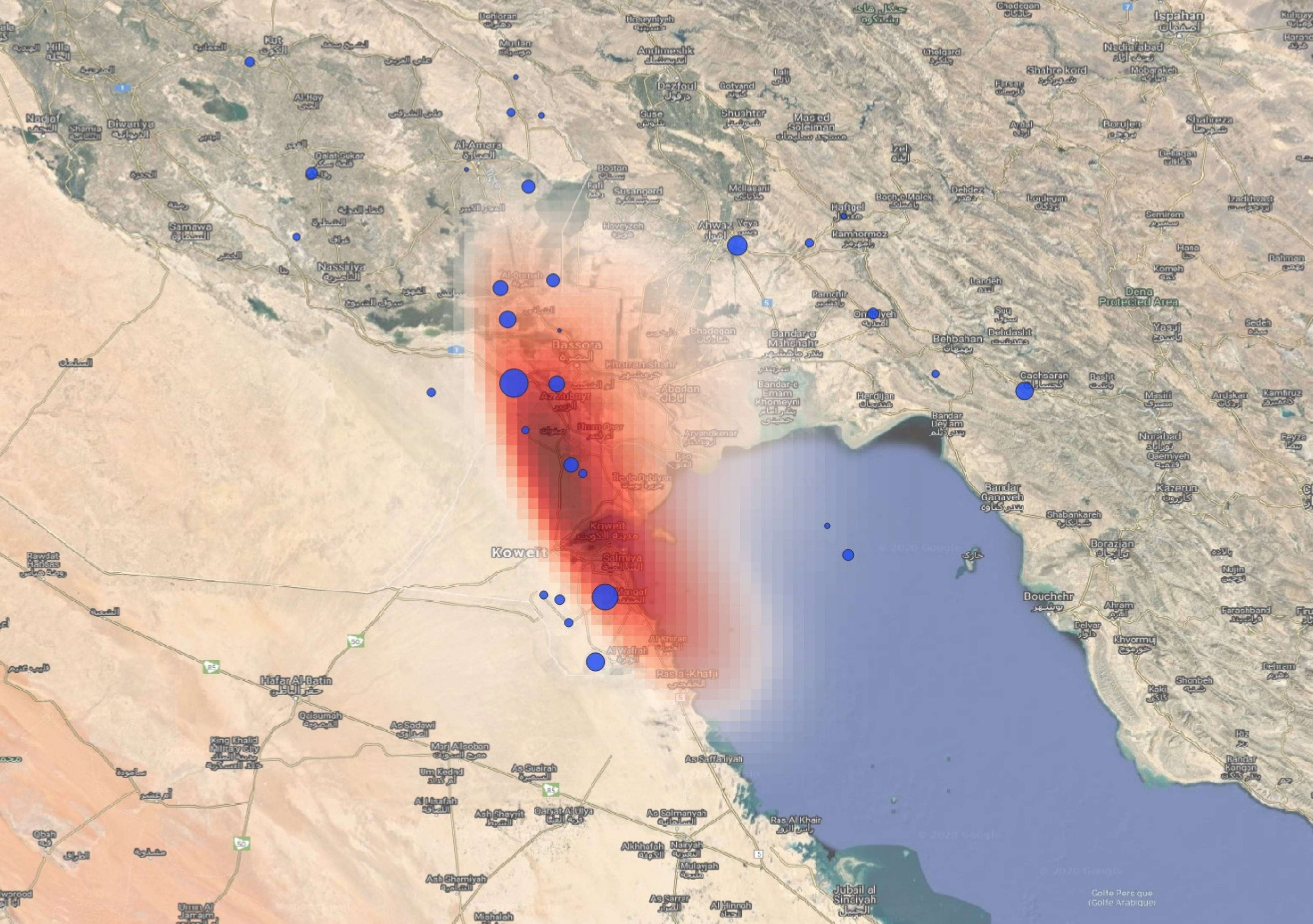}
    \end{tabular}
    \caption{Left: Arithmetic mean. Middle: \eqref{eq:OT-unb-barycenter} + \eqref{eq:l2}. Right: 
    \eqref{eq:OT-unb-barycenter} + \eqref{eq:WFR}}
    \label{fig:irak-bar}
\end{figure}

We also studied the region around the borders between Iraq, Iran and Kuwait, an area known for its high Oil production activity. Figure~\ref{fig:irak-bar} displays the results of our computations. As in the previous example, blue dots correspond the oil fields in the region and the size of the dots represents production level (source: Kayrros analysis). One observes on the averaged image that the satellites measured a high concentration on all the desert areas. This might be due to the well-known albedo-induced bias in the Sentinel-5 Precursor Level 2 Methane data \cite{performancereport}. Using the Wasserstein-Fisher-Rao cost function leads to a clear accumulation of the mass of emissions around the south eastern part of the country, home to the biggest oil field of the country (Rumaila, approximately 1.4 million barrels per day). Due to the high resolution of the image, we couldn't use the \eqref{eq:l2w} metric as it requires too much computation at this point.

\subsection*{Acknowledgements} 
The authors would like to thank Thomas Lauvaux for sharing his expertise on methane production in the Ohio - West Virginia - Pennsylvania region.
AA is at CNRS, and CS Department, Ecole Normale Sup\'erieure, PSL Research University, 45 rue d'Ulm, 75005, Paris and Kayrros SAS. AA would like to acknowledge support from the {\em ML and Optimisation} joint research initiative with the {\em fonds AXA pour la recherche} and Kamet Ventures, a Google focused award, as well as funding by the French government under management of Agence Nationale de la Recherche as part of the "Investissements d'avenir" program, reference ANR-19-P3IA-0001 (PRAIRIE 3IA Institute). Contact: \texttt{aspremon@ens.fr}. MB acknowledges support from an AMX fellowship.

\bibliographystyle{plainnat}
\bibliography{MainPerso,mybiblio}

\appendix
\section{Data Sources}\label{sec:data} 
\subsection{Sentinel 5P}
\begin{figure}[!ht]
\begin{center}
\includegraphics[width=1\columnwidth]{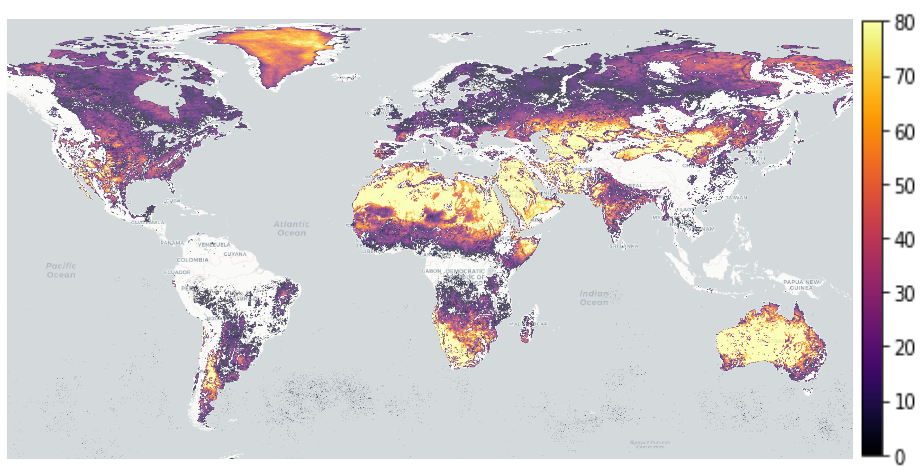} 
    \caption{Sentinel-5P coverage for Level 2 $XCH_4$ data product in 2019. The value of each pixel correspond to the number of days for which Sentinel-5P provided a valid (after quality filtering, \citep{performancereport}) measurement for the corresponding area during year 2019. The data was clipped at 80 for clarity; some pixels exceed this value. \label{fig:coverage2019}}
\end{center}    
\end{figure}

We use total column CH4 (XCH4) measurements from the spaceborne Tropospheric Monitoring Instrument (TROPOMI). TROPOMI is in polar sun‐synchronous orbit and provides global mapping of atmospheric methane columns on daily overpasses at about 13:30 local solar time with 7 $\times$ 7 km nadir pixel resolution (7 $\times$ 5.5 km since June 2019). The mission performance report for Sentinel-5 Precursor Level 2 Methane product \cite{performancereport} states that "the averaged bias for the comparison against 22 TCCON sites is $-0.8\%$ and $-0.31\%$ for the standard and bias corrected XCH4 product". Note that for various reasons (body of water, cloud cover, etc) a significant fraction of the pixels are missing, hence any averaging method used on these images needs to properly account for missing values. Figure \ref{fig:coverage2019} illustrates the variability of the data coverage worldwide.

\subsection{Weather}

Weather data are provided by the ECMWF-ERA5 product at a resolution of 0.25 degree around the equator and sample every hour. We use the northern and eastern direction of the wind measured at 100m above ground to get the direction and the intensity.


\subsection{Production Sites}\label{app:source-prod}
\begin{itemize}
    \item \textbf{Ohio - West Virginia - Pennsylvania:} mines locations can be found  at \url{https://www.pasda.psu.edu/uci/DataSummary.aspx?dataset=257}.
    \item \textbf{Iraq-Iran-Kuweit:} Oil and Gas basins data from Kayrros
    \item \textbf{Permian basin:} completion data from Kayrros
\end{itemize}

\subsection{Maps}
We used google earth for the maps underlying our plots.

\section{Toy Example}\label{app:toy-wind}

\begin{figure}[!ht]
    \centering
    \includegraphics[width=0.6\linewidth]{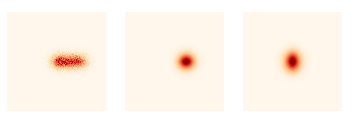}
    \caption{From left to right: arithmetic mean of the $g^{(k)}$, Wasserstein barycenter of the $g^{(k)}$ with \eqref{eq:l2} cost, Wasserstein barycenter of the $g^{(k)}$ with \eqref{eq:l2w} cost using constant East wind. }
    \label{fig:wind-gaussian}
\end{figure}

This toy example consists in choosing the $g^{(k)}$ as Gaussian clouds of unit variance and mean $\mu_k$ that drift away from the center to the right of the image. In Figure~\ref{fig:wind-gaussian} we compare the results of Wasserstein barycenters with different cost matrix. For the right plot we consider the \eqref{eq:l2w} cost with a constant wind from the left to the right. Thus the barycenter has been translated in the opposite direction of the wind. 

\section{Wind in Pennsylvania}\label{app:wind-pen}
\begin{figure}[!ht]
    \centering
    \includegraphics[scale=0.2]{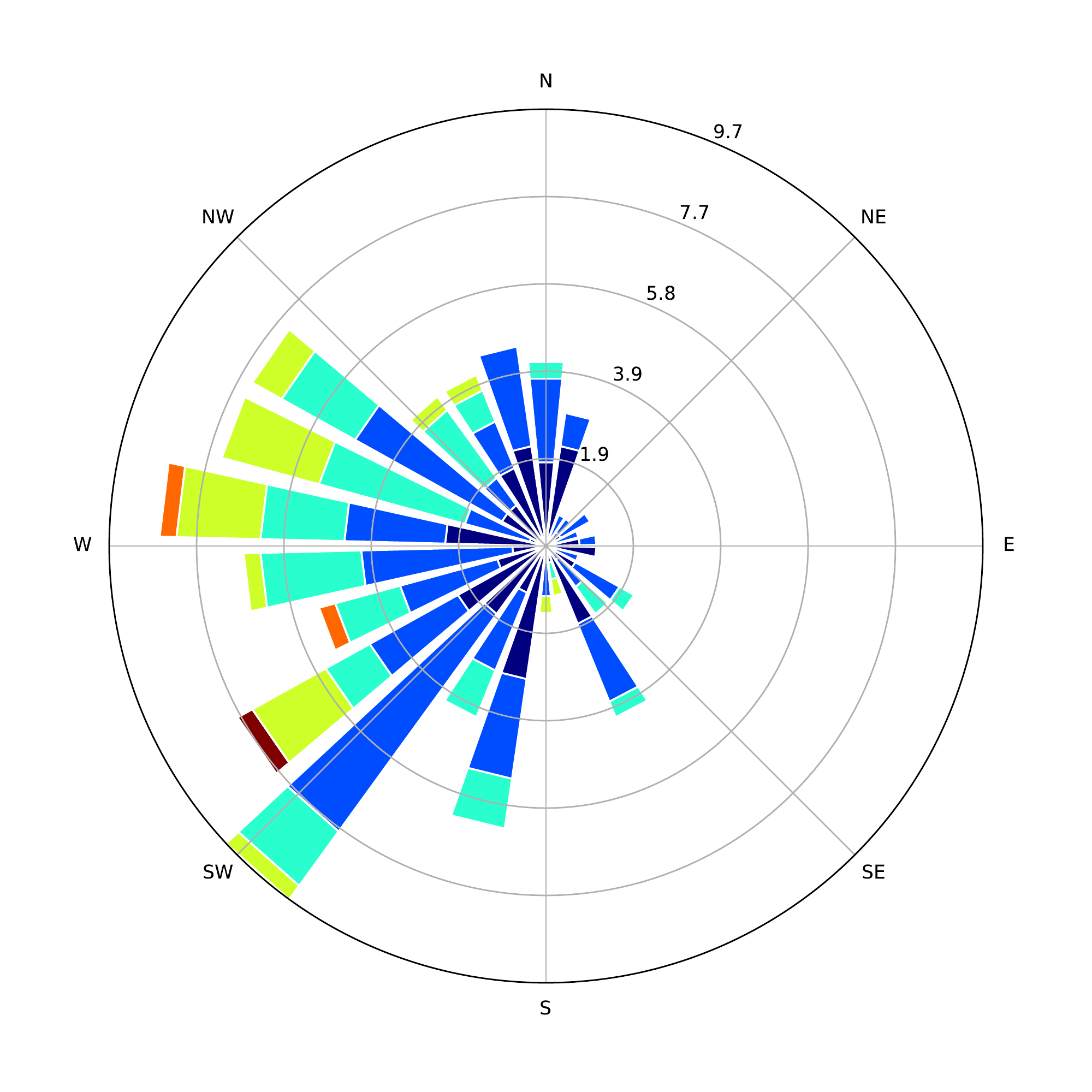}
    \caption{Distribution of the mean wind speed and direction on the Pennsylvania mining region.}
    \label{fig:pen-windrose}
\end{figure}

Figure~\ref{fig:pen-windrose} represents the distribution of the wind mean directions and speeds during the studied period. Bars indicates the direction from which the wind comes from (meteorological convention), the size of the bars corresponds to the directions frequencies and the color to the speed (blue is slow and red is fast). We notice that in this region the wind is meanly blowing on some West-East direction.

\section{Permian}\label{app:permian}

\begin{figure}[!ht]
    \centering
    \begin{tabular}{cc}
\includegraphics[width=0.45\linewidth]{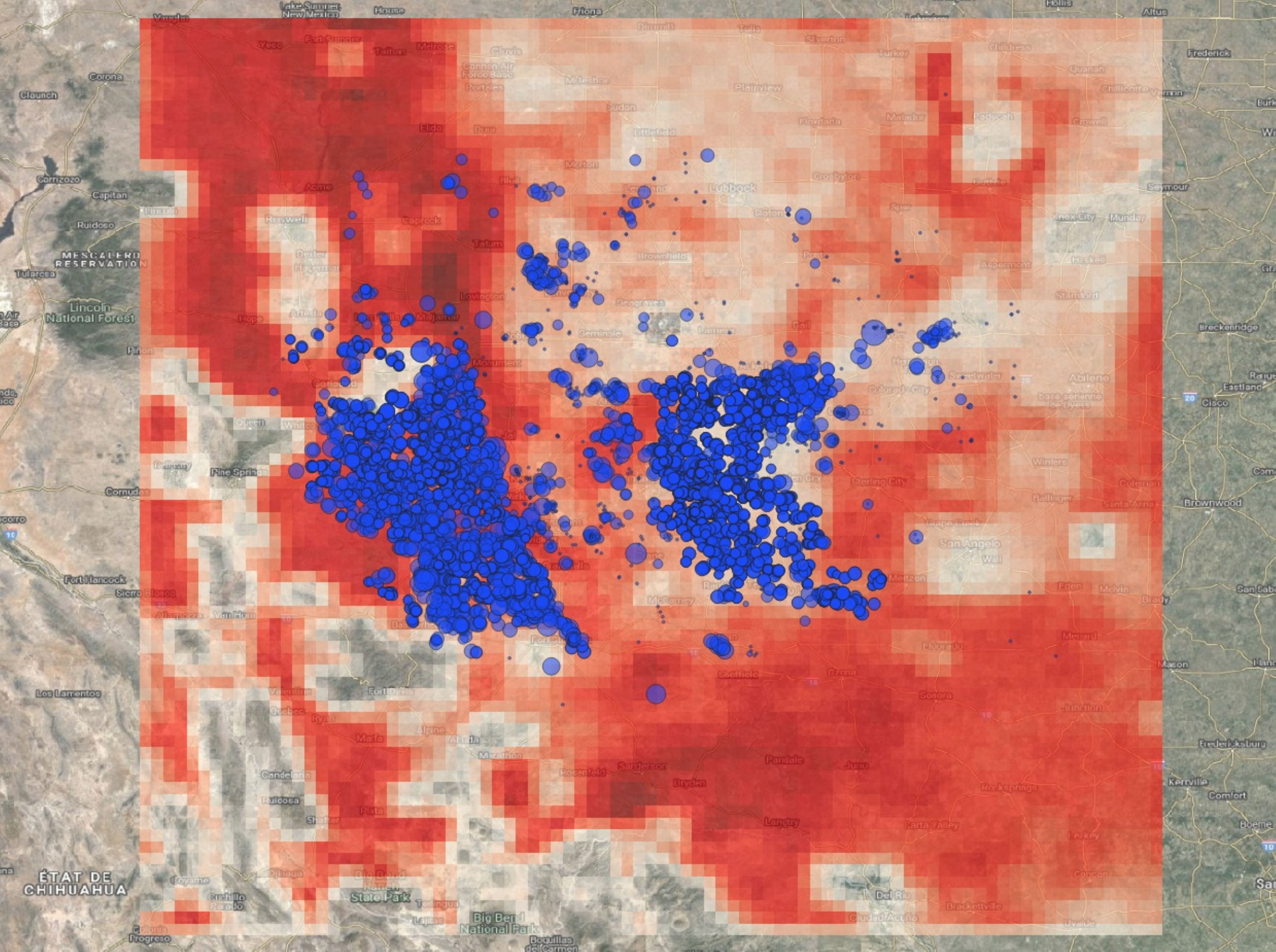}&\hspace{-0.3cm}\includegraphics[width=0.45\linewidth]{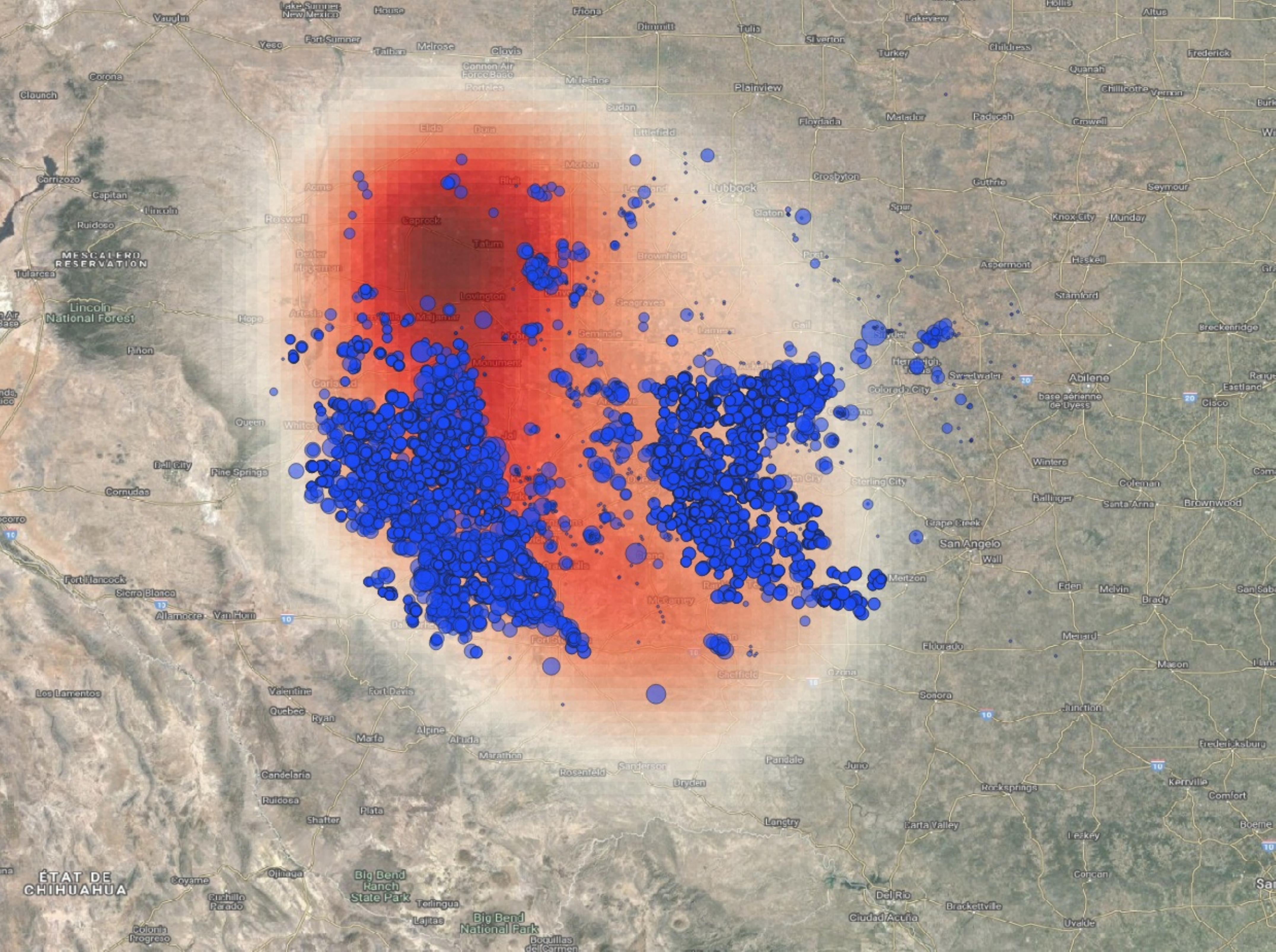}\\
\includegraphics[width=0.45\linewidth]{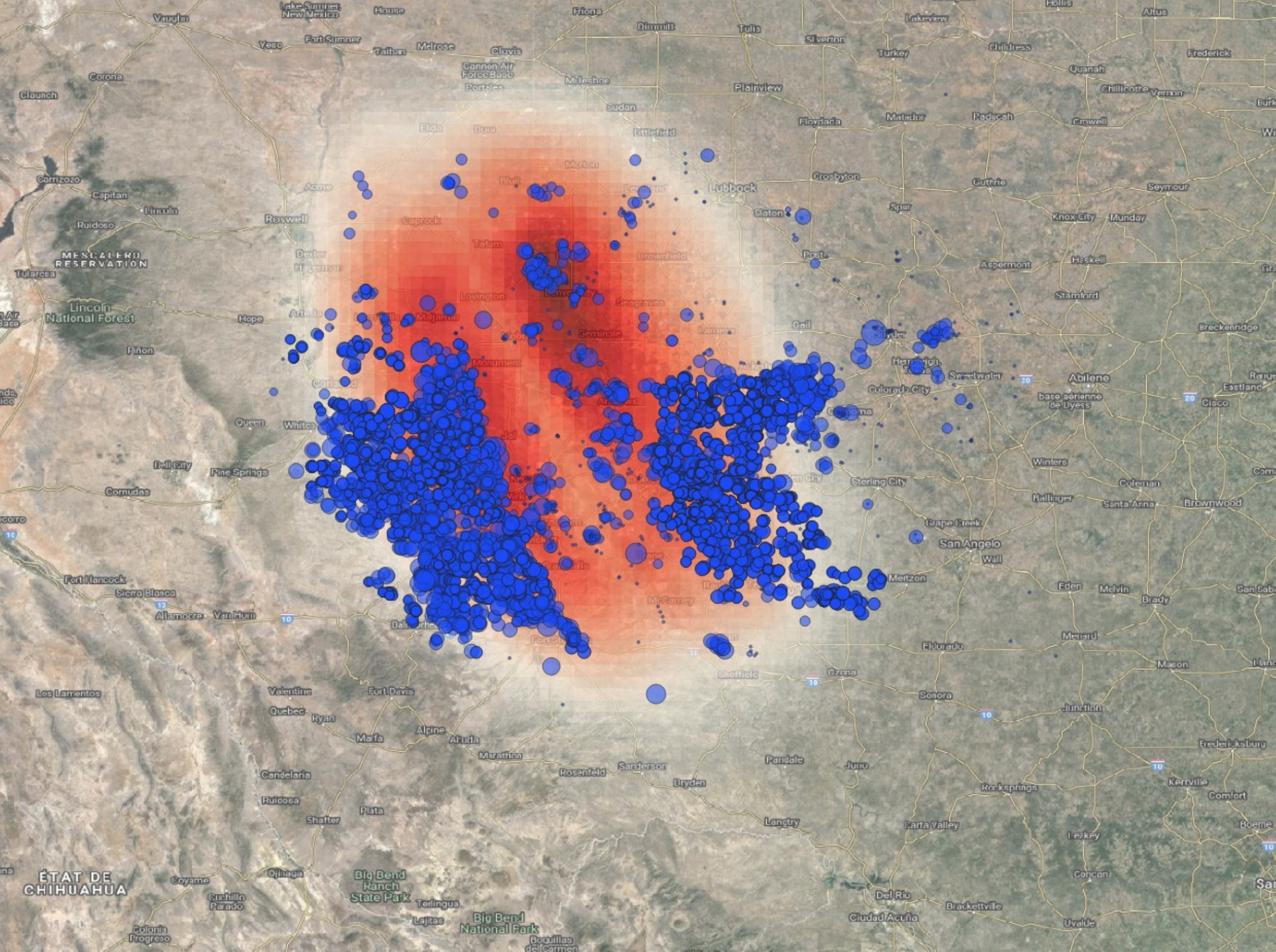}&\hspace{-0.3cm}\includegraphics[width=0.45\linewidth]{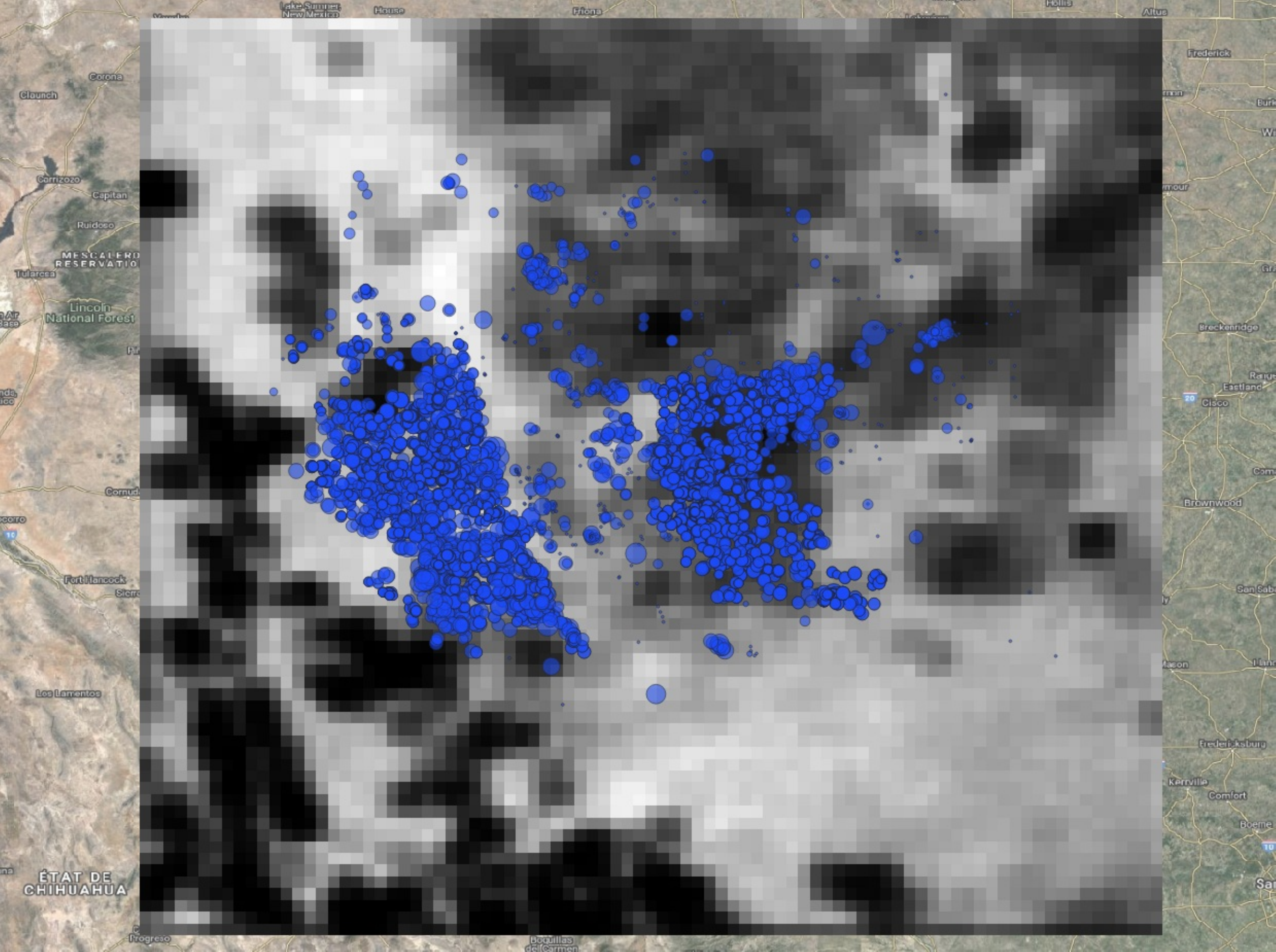}
    \end{tabular}
    \caption{Top Left: Arithmetic mean. Top Right: \eqref{eq:OT-unb-barycenter} + \eqref{eq:l2}. Bottom Left: 
    \eqref{eq:OT-unb-barycenter} + \eqref{eq:WFR}}. Bottom Right: Proportion of observed pixels (black is 0\%, white is 30\%).
    \label{fig:permian-bar}
\end{figure}

In this section we focus on the Permian basin. This area is known for its high production of shale Oil and Gas. The result of our experiments on this region is displayed on Figure~\ref{fig:permian-bar}.
The blue dots represent the locations of the recently completed oil wells (source: Kayrros analysis), over a period where production reached its maximum capacity. The dot size is proportional with the quantity of Oil and Gas produced by the well. 

One can observe on Figure~\ref{fig:permian-bar} Bottom Right that the quantity of missing observations in the satellite data is quite high. In particular, we get very few observations in a region with high source density on the East side. However Wasserstein barycenter gives some mass to this region, and using Wasserstein-Fisher-Rao metric even allows to separate the two production sub basins. These two areas correspond to the two major oil and gas sub-basins within the Permian, namely the Delaware basin and the Midland basin.

Due to the higher resolution of these observations, we couldn't use the metric involving the wind data as it became too computationally intensive.

\end{document}